\pdfoutput=1

\documentclass[11pt,a4paper]{article}
\usepackage[hyperref]{acl2019}

\usepackage{algorithmicx, algorithm, algpseudocode}
\usepackage{times}
\usepackage{booktabs}
\usepackage{latexsym}
\usepackage{float}
\usepackage{hyperref}
\usepackage{graphicx}
\usepackage{makecell}
\usepackage{wrapfig}
\usepackage{url}
\usepackage[normalem]{ulem}

\def\BaseModel{FactCC}
\def\ExplainableModel{FactCCX}
\def\PositiveLabel{\texttt{CONSISTENT}}
\def\NegativeLabel{\texttt{INCONSISTENT}}

\newcommand{\DataGenerationAlgorithm}{
    \begin{figure}
    \small
    \begin{algorithmic}[0]
        \Require{
        \\S - set of source documents 
        \\ $\mathcal{T}^+$ - set of semantically invariant transformations
        \\ $\mathcal{T}^-$ - set of semantically variant transformations}
        \State
        \Function{$\texttt{generate\_data}$}{$\mathcal{S}$, $\mathcal{T}^+$, $\mathcal{T}^-$}
    	\State $\mathcal{D} \gets \emptyset$ 
    	\Comment set of generated data points
    	
    	\For{$doc \textbf{ in } \mathcal{S}$}
    		\State $doc\_sents \gets \texttt{sentence\_tokenizer}(doc)$
    		\State $sent \gets \texttt{choose\_random}(doc\_sents)$
    		\State $\mathcal{D} \gets \mathcal{D} \cup \{ (doc, sent, +) \}$
    		\For{$\texttt{fn} \textbf{ in } \mathcal{T}^+$}
    		    \State $new\_sent \gets \texttt{fn}(doc, sent)$
        		\State $\mathcal{D} \gets \mathcal{D} \cup \{ (doc, new\_sent, +) \}$
            \EndFor
       	\EndFor         
       	
       	\State
    	\For{$example \textbf{ in } \mathcal{D}$}
    	    \State $doc, sent, \_ \gets example$
    		\For{$\texttt{fn} \textbf{ in } \mathcal{T}^-$}
    		    \State $new\_sent \gets \texttt{fn}(doc, sent)$
        		\State $\mathcal{D} \gets \mathcal{D} \cup \{ (doc, new\_sent, -) \}$
            \EndFor
       	\EndFor         
    	\State \Return $\mathcal{D}$
        \EndFunction
    \end{algorithmic}
    \caption{
    Procedure to generate weakly-supervised training data.
    $\mathcal{S}$ is a set of source documents,
    $\mathcal{T^+}$ is a set of semantically invariant text transformations,
    $\mathcal{T^-}$ is a set of semantically variant text transformations,
    $+$ is a positive label, 
    $-$ is a negative label.
    }
    \label{fig:data-generation-code}
    \end{figure}
}

\newcommand{\SummarizationErrorsTable}{
    \begin{table*}[th]
        \begin{center}
        \resizebox{\linewidth}{!}{%
        \small
        \begin{tabular}{p{0.48\linewidth}|p{0.48\linewidth}} 
        \toprule
        \multicolumn{2}{l}{\textbf{Source article fragments}} \\ 
        \midrule
        (CNN) \textcolor{teal}{The mother of a quadriplegic man who police say was left in the woods for days cannot be extradited} to face charges in Philadelphia until she completes an unspecified "treatment," Maryland police said Monday. The Montgomery County (Maryland) Department of Police took \textcolor{teal}{Nyia Parler, 41}, into custody Sunday (...) &
        (CNN) The classic video game "Space Invaders" was developed in Japan \textcolor{teal}{back in the late 1970's} -- and now their real-life counterparts are the topic of an earnest political discussion in Japan's corridors of power. Luckily, Japanese can sleep soundly in their beds tonight as the government's top military official earnestly revealed that (...) \\
        \midrule
        \multicolumn{2}{l}{\textbf{Model generated claims}} \\
        \midrule
        \textcolor{red}{Quadriplegic man} Nyia Parler, 41, \textcolor{red}{left in woods for days} can not be extradited. &
        Video game "Space Invaders" was developed in Japan back \textcolor{red}{in 1970}. \\
        \bottomrule
        \end{tabular}
        }%
        \caption{
        Examples of factually incorrect claims output by summarization models.
        Green text highlights the support in the source documents for the generated claims, red text highlights the errors made by summarization models.
        }
        \label{tab:summarization-errors}
        \end{center}
    \end{table*}
}

\newcommand{\DataTransformationTable}{
    \begin{table*}[t!]
        \begin{center}
        \resizebox{\linewidth}{!}{%
            \small
            \begin{tabular}{p{0.16\linewidth}|p{0.40\linewidth} p{0.40\linewidth}} 
            \toprule
            \textbf{Transformation} & \textbf{Original sentence} & \textbf{Transformed sentence} \\
            \midrule
            Paraphrasing        & 
            \textcolor{teal}{Sheriff Lee Baca has now decided to recall some 200 badges his department has handed out to local politicians just two weeks after the picture was released by the U.S. attorney's office in support of bribery charges against three city officials.} &
            \textcolor{red}{Two weeks after the US Attorney's Office issued photos to support bribery allegations against three municipal officials, Lee Baca has now decided to recall about 200 badges issued by his department to local politicians.} \\
            \midrule
            
            Sentence negation   &
            Snow \textcolor{teal}{was} predicted later in the weekend for Atlanta and areas even further south. & 
            Snow \textcolor{red}{wasn't} predicted later in the weekend for Atlanta and areas even further south. \\
            \midrule
            
            Pronoun swap        & 
            It comes after \textcolor{teal}{his} estranged wife Mona Dotcom filed a \$20 million legal claim for cash and assets. & 
            It comes after \textcolor{red}{your} estranged wife Mona Dotcom filed a \$20 million legal claim for cash and assets. \\
            \midrule
            
            Entity swap         & 
            Charlton coach \textcolor{teal}{Guy Luzon} had said on Monday: 'Alou Diarra is training with us.' &
            Charlton coach \textcolor{red}{Bordeaux} had said on Monday: 'Alou Diarra is training with us.' \\
            \midrule
            
            Number swap         & 
            He says he wants to pay off the \textcolor{teal}{\$12.6million} lien so he can sell the house and be done with it, according to the Orlando Sentinel. &
            He says he wants to pay off the \textcolor{red}{\$3.45million} lien so he can sell the house and be done done with it, according to the Orlando Sentinel. \\
            \midrule
            
            Noise injection   &
            Snow \textcolor{teal}{was} predicted later in the weekend for Atlanta and areas even further south. & 
            Snow \textcolor{teal}{was was} predicted later in the weekend for Atlanta and areas \textcolor{red}{\sout{even}} further south. \\
            \midrule
            \bottomrule
            \end{tabular}
        }%
        \caption{
        Examples of text transformations used to generate training data.
        Green and red text highlight the changes made by the transformation.
        Paraphrasing is a semantically invariant transformation, 
        Sentence negation, entity, pronoun, and number swaps are semantically variant transformation.
        }
        \label{tab:data-transformations}
        \end{center}
    \end{table*}
}

\newcommand{\ModelPerformanceTable}{
    \begin{table}[t!]
        \begin{center}
        \begin{tabular}{l|cc} 
        \toprule
        Model & \makecell{Accuracy \\ \textit{(weighted)}} & F1-score \\
        \midrule
        BERT+MNLI        & 51.51 & 0.0882 \\
        BERT+FEVER       & 52.07 & 0.0857 \\
        \midrule
        \BaseModel{} (ours)     & \textbf{74.15} & \textbf{0.5106} \\
        \ExplainableModel{} (ours)   & 72.88 & 0.5005 \\
        \bottomrule
        \end{tabular}
        \caption{
        Performance of models evaluated by means of weighted (class-balanced) accuracy and F1 score on the manually annotated test set.
        }
        \label{tab:model-performance}
        \end{center}
    \end{table}
}

\newcommand{\SentencePairPerformanceTable}{
    \begin{table}[t!]
        \begin{center}
        \resizebox{\linewidth}{!}{%
            \small
            \begin{tabular}{l|cc} 
            \toprule
            Model & Incorrect & $\Delta$ \\
            \midrule
            Random                      & 50.0\%    & \\
            \midrule
            DA~\cite{Falke:19}          & 42.6\%    & -7.4  \\
            InferSent~\cite{Falke:19}   & 41.3\%    & -8.7  \\
            SSE~\cite{Falke:19}         & 37.3\%    & -12.7 \\
            BERT~\cite{Falke:19}        & 35.9\%    & -14.1 \\
            ESIM~\cite{Falke:19}        & 32.4\%    & -17.6 \\
            \midrule
            \BaseModel{} (ours)                & \textbf{30.0\%} & \textbf{-20.0} \\
            \bottomrule
            \end{tabular}
        }%
        \caption{
        Percentage of incorrectly ordered sentence pairs using different consistency prediction mnodels and crowdsourced human performance on the dataset.
        }
        \label{tab:sentence-pair-performance}
        \end{center}
    \end{table}
}

\newcommand{\ExplainableOutputsTable}{
    \begin{table*}[th]
        \begin{center}
        \resizebox{\linewidth}{!}{%
        \small
        \begin{tabular}{p{0.98\linewidth}} 
        \toprule
        \textbf{Article} \\ 
        \midrule
        (CNN) Blues legend B.B. King was hospitalized for dehydration, though the ailment didn't keep him out for long. King's dehydration was caused by his Type II diabetes, but he "is much better," his daughter, Claudette King, told the Los Angeles Times. The legendary guitarist and vocalist released a statement thanking those who have expressed their concerns. "I'm feeling much better and am leaving the hospital today," King said in a message Tuesday. \textcolor{orange}{Angela Moore, a publicist for Claudette King, said later in the day that he was back home resting and enjoying time with his grandchildren.} "He was struggling before, and he is a trouper," Moore said. "He wasn't going to let his fans down." No more information on King's condition or where he was hospitalized was immediately available. (...)
        \\
        \midrule
        \textbf{Claim} \\
        \midrule
        \textcolor{red}{Angela Moore} was back home resting and enjoying time with his grandchildren. \\
        \bottomrule
        \end{tabular}
        }%
        \caption{
        Example of a test pair correctly classified as \textit{incorrect} and highlighted by our explainable model.
        Orange text indicates the span of the source documents that should contain support for the claim.
        Red text indicates the span of the claim that was selected as incorrect.
        }
        \label{tab:explainable-predictions}
        \end{center}
    \end{table*}
}

\newcommand{\MturkResultsTable}{
    \begin{table*}[th!]
        \begin{center}
        \begin{tabular}{r|ccc|cc} 
        & \multicolumn{3}{c|}{Model Highlight Helpfulness} & \multicolumn{2}{c}{\makecell{Model-Annotator \\ Highlight Overlap}} \\
        \midrule
        Annotation subset & Helpful & Somewhat Helpful & Not Helpful & Accuracy & F1 score \\
        \midrule
        \multicolumn{6}{c}{\textit{Article Highlights}} \\
        \midrule
        Raw Data            & $79.21\%$ & $12.54\%$ & $8.25\%$ & $65.33\%$ & $0.6207$ \\
        Golden Aligned      & $77.73\%$ & $12.66\%$ & $9.61\%$ & $74.87\%$ & $0.7161$ \\
        Majority Aligned    & $81.11\%$ & $11.48\%$ & $7.41\%$ & $69.88\%$ & $0.6679$ \\
        \midrule
        \multicolumn{6}{c}{\textit{Claim Highlights}} \\
        \midrule
        Raw Data            & $64.44\%$ & $16.89\%$ & $18.67\%$ & $65.66\%$ & $0.6650$ \\
        Golden Aligned      & $67.28\%$ & $16.05\%$ & $16.67\%$ & $80.54\%$ & $0.8190$ \\
        Majority Aligned    & $67.17\%$ & $16.67\%$ & $16.16\%$ & $69.48\%$ & $0.6992$ \\
        \bottomrule
        \end{tabular}
        \caption{
        Quality of spans highlighted in the \textit{article} and \textit{claim} by the \ExplainableModel{} model evaluated by human annotators.
        The left side shows whether the highlights were considered helpful for the task of factual consistency annotations.
        The right side shows the overlap between model generated and human annotated highlights.
        Different rows show how the scores change depending on how the collected annotations are filtered.
        \textit{Raw Data} shows results without filtering, \textit{Golden Aligned} only considers annotations where the human-assigned label agreed with the author-assigned label, \textit{Majority Aligned} only considers annotations where the human-assigned label agreed with the majority-vote label from all annotators.
        }
        \label{tab:mturk-results}
        \end{center}
    \end{table*}
}

\newcommand{\HighlightImprovementsTable}{
    \begin{table}[t!]
        \begin{center}
        \resizebox{\linewidth}{!}{%
            \small
            \begin{tabular}{l|c|c} 
            \toprule
            & \makecell{Task without \\ model highlights} & \makecell{Task with \\ model highlights} \\
            \midrule
            \makecell{Average work \\ time (sec)} & $224.89$ & $178.34$ \\
            \makecell{Inter-annotator \\ agreement ($\kappa$)} & $0.1571$ & $0.2526$ \\
            \bottomrule
            \end{tabular}
        }%
        \caption{
        Annotation speed and inter-annotator agreement measured for factual consistency checking with and without assisting, model generated highlights.
        }
        \label{tab:highlight-improvements}
        \end{center}
    \end{table}
}

\aclfinalcopy 

\title{Evaluating the Factual Consistency of Abstractive Text Summarization}

\author{Wojciech Kry\'sci\'nski, Bryan McCann, Caiming Xiong, Richard Socher \\
        Salesforce Research \\
        \texttt{\{kryscinski,bmccann,cxiong,rsocher\}@salesforce.com} \\}

\date{}

\begin{document}
\maketitle
\begin{abstract}
Currently used metrics for assessing summarization algorithms do not account for 
whether summaries are factually consistent with source documents.
We propose a weakly-supervised, model-based approach for verifying factual consistency and identifying conflicts between source documents and a generated summary.
Training data is generated by applying a series of rule-based transformations to the sentences of source documents. 
The factual consistency model is then trained jointly for three tasks: 1) identify whether sentences remain factually consistent after transformation, 2) extract a span in the source documents to support the consistency prediction, 3) extract a span in the summary sentence that is inconsistent if one exists.
Transferring this model to summaries generated by several state-of-the art models reveals that this highly scalable approach substantially outperforms previous models, including those trained with strong supervision using standard datasets for natural language inference and fact checking.
Additionally, human evaluation shows that the auxiliary span extraction tasks provide useful assistance in the process of verifying factual consistency.
\end{abstract}
 
\section{Introduction}
\SummarizationErrorsTable
The goal of text summarization models is to transduce long documents into a shorter form that retains the most important aspects of the source document.
Common approaches to summarization are \textit{extractive}~\citep{Dorr:03, Nallapati:17} where the model directly copies the salient parts of the source document into the summary, \textit{abstractive}~\citep{Rush:15, Paulus:17} where the important parts are paraphrased to form novel sentences, and \textit{hybrid}~\citep{Gehrmann:18, Hsu:18, Chen:18}, combining the two methods by employing specialized extractive and abstractive components.

Advancements in neural architectures~\citep{Cho:14, Sutskever:14, Bahdanau:14, Vinyals:15, Vaswani:17}, pre-training and transfer learning \cite{McCann:17, Peters:2018, Devlin:18}, and availability of large-scale supervised datasets~\citep{Sandhaus:08, Nallapati:16, Grusky:18, Narayan:18, Sharma:19} allowed deep learning-based approaches to dominate the field. State-of-the-art solutions utilize self-attentive Transformer blocks~\citep{YLiu:19a, YLiu:19b, Zhang:19}, attention and copying mechanisms~\citep{See:17, Cohan:18}, and multi-objective training strategies~\citep{Guo:18, Pasunuru:18}, including reinforcement learning techniques~\citep{Kryscinski:18, Dong:18, Wu:18}.

Despite significant efforts made by the research community, there are still many challenges limiting progress in summarization: 
insufficient evaluation protocols that leave important dimensions, such as factual consistency, unchecked, 
noisy, automatically collected datasets that leave the task underconstrained, 
and strong, domain-specific layout biases in the data that dominate training signal~\citep{Kryscinski:19}.

We address the problem of verifying factual consistency
between source documents and generated summaries:
a factually consistent summary contains only statements that are entailed by the source document.
Recent studies show that up to 30\% of summaries generated by abstractive models contain factual inconsistencies~\citep{Cao:18, Goodrich:19, Falke:19, Kryscinski:19}.
Such high levels of factual inconsistency render automatically generated summaries virtually useless in practice.

The problem of factual consistency is closely related to natural language inference (NLI) and fact checking.
Current NLI datasets~\citep{Bowman:15, Conneau:18, Williams:18} focus on classifying logical entailment between short, single sentence pairs, but verifying factual consistency can require incorporating the entire context of the source document.
Fact checking focuses on verifying facts against the whole of available knowledge, whereas factual consistency checking focuses on adherence of facts to information provided by a source document without guarantee that the information is true.

We propose a novel, weakly-supervised BERT-based~\citep{Devlin:18} model for verifying factual consistency, and we add specialized modules that explain which portions of both the source document and generated summary are pertinent to the model's decision.
Training data is generated from source documents by applying a series of rule-based transformations that were inspired by error-analysis of state-of-the-art summarization model outputs.
Training with this weak supervision substantially improves over using the strong supervision provided by existing datasets for NLI~\citep{Williams:18} and fact-checking~\citep{Thorne:18}.
Through human evaluation we show that the explanatory modules that augment our factual consistency model provide useful assistance to humans as they verify the factual consistency between a source document and generated summaries.
\section{Related Work}
This work builds on prior work for factual consistency in text summarization and natural language generation.
\citet{Goodrich:19} propose an automatic, model-dependent metric for evaluating the factual accuracy of generated text.
Facts are represented as \textit{subject-relation-object} triplets and factual accuracy is defined as the precision between facts extracted from the generated summary and source document.
The authors proposed a new dataset for training fact extraction models based on Wikipedia articles and used it to train a Transformer-based architecture to extract fact triplets.
Factual accuracy was then measured by applying this model to the outputs of a separate text summarization model, which had been trained to generate the introduction sections of Wikipedia articles from a set of reference documents~\cite{PLiu:18}.
Human evaluation demonstrated that the proposed technique outperformed other, non-model based, evaluation metrics, such as ROUGE~\cite{Lin:04}, in assessing factual accuracy.
Despite positive results, the authors highlighted remaining challenges, such as its inability to adapt to negated relations or relation names expressed by synonyms.

A parallel line of research focused on improving factual consistency of summarization models by exploring different architectural choices and strategies for both training and inference.
In~\citet{Falke:19}, the authors proposed re-ranking potential potential summaries based on factual correctness during beam search.
The solution used textual entailment (NLI) models, trained on the SNLI~\cite{Bowman:15} and MNLI~\citep{Williams:18} datasets, to score summaries by means of the entailment probability between all source document-summary sentence pairs.
The summary with the highest aggregate entailment score was used as the final output of the summarization model.
The authors validated their approach using summaries generated by models trained on the CNN/DailyMail dataset~\citep{Nallapati:16}.
The authors concluded that out-of-the-box NLI models do not transfer well to the task of factual correctness.
The work also showed that the ROUGE metric~\citep{Lin:04}, commonly used to evaluate summarization models, does not correlate with factual correctness.
In~\citet{Cao:18}, the authors proposed a novel, dual-encoder architecture that in parallel encodes the source documents and all the facts contained in them.
During generation, the decoder attends to both the encoded source and facts which, according to the authors, forces the output to be conditioned on the both inputs.
The facts encoded by the model are explicitly extracted by out-of-the-box open information extraction and a dependency parsing models.
Experiments were conducted on the Gigaword~\citep{Graff:03} dataset, through human evaluation the authors showed that the proposed technique substantially lowered the number of errors in generated single-sentence summaries.
\citet{HLi:18} incorporated entailment knowledge into summarization by introducing an entailment-aware encoder-decoder model for sentence summarization.
Entailment knowledge is injected in two ways: the encoder is shared for the task of summarization and textual entailment and the decoder is trained using reward augmented maximum likelihood (RAML) with rewards coming from a pre-trained entailment classifier.
Experiments conducted on the Gigaword~\citep{Graff:03} dataset showed improvements against baselines on both correctness and informativeness.

More loosely related work explored training summarization models in multi-task~\citep{Guo:18} and multi-reward~\citep{Pasunuru:18} settings where the additional task and reward was textual entailment (NLI).
The intuition was that incorporating NLI in the training procedure should improve entailment between the summary and source document, however, neither of the mentioned works conducted studies or analysis that would verify this.

\DataTransformationTable

\section{Methods}\label{sec:methods}
A careful study of the outputs of state-of-the-art summarization models provided us with valuable insights about the specifics of factual errors made during generation and possible means of detecting them.
Primarily, checking factual consistency on a \textit{sentence-sentence} level, where each sentence of the summary is verified against each sentence from the source document, is insufficient.
Some cases might require a longer, multi-sentence context from the source document due to ambiguities present in either of the compared sentences.
Summary sentences might paraphrase multiple fragments of the source document, while source document sentences might use certain linguistic constructs, such as coreference, which bind different parts of the document together.
In addition, errors made by summarization models are most often related to the use of incorrect entity names, numbers, and pronouns.
Other errors such as negations and common sense error occur less often.~\footnote{A more fine-grained taxonomy of errors could be created, where, for example, incorrectly attributing quotes to entities would be distinguished from choosing an incorrect subject in a sentence. However, it would carry the implicit assumption that NLP models have the ability to reason about the processed text in a similar way as humans do. We refrain from anthropomorphizing summarization models.}
Taking these insights into account, we propose and test a \textit{document-sentence} approach for factual consistency checking, where each sentence of the summary is verified against the entire body of the source document.

\subsection{Training data}\label{sec:generated-data}
Currently, there are no supervised training datasets for factual consistency checking.
Creating a large-scale, high-quality dataset with strong supervision collected from human annotators is prohibitively expensive and time consuming. 
Thus alternative approaches of acquiring training data are necessary.

Considering the current state of summarization, in which the level of abstraction of generated summaries is low and models mostly paraphrase single sentences and short spans from the source~\citep{Kryscinski:18, Zhang:18}, we propose using an artificial, weakly-supervised dataset for the task at hand.
Our data creation method requires an unannotated collection of source documents in the same domain as the summarization models that are to be checked.
Examples are created by first sampling single sentences, later referred to as \textit{claims}, from the source documents.
Claims then pass through a set of textual transformations that output novel sentences with both \textit{positive} and \textit{negative} labels.
A detailed description of the data generation function is presented in Figure~\ref{fig:data-generation-code}.
The obvious benefit of using an artificially generated dataset is that it allows for creation of large volumes of data at a marginal cost.
The data generation process also allows to collect additional metadata that can be used in the training process.
In our case, the metadata contains information about the original location of the extracted claim in the source document and the locations in the claim where text transformations were applied.

Our data generation process incorporates both semantically invariant ($\mathcal{T^+}$), and variant ($\mathcal{T^-}$) text transformations to generate novel claims with \texttt{CORRECT} and \texttt{INCORRECT} labels accordingly.
%
This work uses the following transformations:

\paragraph{Paraphrasing}
A paraphrasing transformation covers cases where source document sentences are rephrased by the summarization model.
%
Paraphrases were produced by backtranslation using Neural Machine Translation systems~\citep{Edunov:18}.
The original sentence was translated to an intermediate language and translated back to English yielding a semantically-equivalent sentence with minor syntactic and lexical changes.
\textit{French}, \textit{german}, \textit{chinese}, \textit{spanish}, and \textit{russian} were used as intermediate languages.
These languages were chosen based on the performance of recent NMT systems with the expectation that well-performing languages could ensure better translation quality.
We used the Google Cloud Translation API~\footnote{\url{https://cloud.google.com/translate/}} for translations.

\paragraph{Entity and Number swapping}
To learn how to identify examples where the summarization model uses incorrect numbers and entities in generated text we used the Entity and Number swapping transformation.
An NER system was applied to both the claim sentence and source document to extract all mentioned entities.
To generate a novel, semantically changed claim, an entity in the claim sentence was replaced with an entity from the source document.
Both of the swapped entities were chosen at random while ensuring that they were unique.
Extracted entities were divided into two groups, named entities, covering person, location and institution names, and number entities, such as dates and all other numeric values.
Entities were swapped within their groups, i.e. names entities would only be replaced with other named entities.
In this work we used the we used the SpaCy NER tagger~\citep{Honnibal:17}.

\paragraph{Pronoun swapping}
To teach the factual consistency checking model how to find incorrect pronoun use in claim sentences we used a pronoun swapping data augmentation.
All gender-specific pronouns were first extracted from the claim sentence.
Next, a randomly chosen pronoun was swapped with a different one from the same pronoun group to ensure syntactic correctness, i.e. a possessive pronoun could only be replaced with another possessive pronoun.
New sentences were considered semantically variant.

\paragraph{Sentence negation}
To give the factual consistency checking model the ability to handle negated sentences we used a sentence negation transformation.
In the first step, claim sentence was scanned in search of auxiliary verbs.
To switch the meaning of the new sentence, a randomly chosen auxiliary verb was replaced with its negation.
Positive sentences would be negated by adding \textit{not} or \textit{n't} after the chosen verb, negative sentences would be switched by removing the negation.

\paragraph{Noise injection}
Because a verified summary is fully generated by a deep neural network, they should be expected to contain certain types of noise.
In order to make the trained factual consistency model robust to such generation errors, all training examples were injected with noise using a simple algorithm.
For each token in a claim the decision was made whether noise should be added at the given position with a preset probability.
If noise should be injected, the token was randomly duplicated or removed from the sequence. 
Examples of all transformations are presented in Table~\ref{tab:data-transformations}.

\subsection{Development and test data} \label{sec:manual-data}
Apart from the artificially generated training set, separate, manually annotated, development and test sets were created.
Both of the manually annotated dataset utilized summaries output by state-of-the-art summarization models.
Each summary was split into separate sentences and all \textit{(document, sentence)} pairs and annotated by the authors of this work.
Since the focus was to collect data that would allow to verify the factual consistency of summarization models, any unreadable sentences caused by poor generation were not labeled.
The development set consists of 931 examples, the test set contains 503 examples.
The model outputs used for annotation were provided by the authors of papers:~\citet{Hsu:18, Gehrmann:18, Jiang:18, Chen:18, See:17, Kryscinski:18, WLi:18, Pasunuru:18, Zhang:18, Guo:18}.

Effort was made to collect a larger set of annotations through crowdsourcing platforms, however the inter-annotator agreement and general quality of annotations was too low to be considered reliable for the task at hand. 
This aligns with the conclusions of \cite{Falke:19}, where the authors showed that for the task of factual consistency the inter-annotator agreement coefficient $\kappa$ reached 0.75 only when 12 annotations were collected for each example.
This in turn yields prohibitively high annotations costs.

\DataGenerationAlgorithm
\subsection{Models}
Considering the significant recent improvements in natural language understanding (NLU) tasks (including natural language inference) coming from using pre-trained Transformer-based models~\footnote{\url{http://gluebenchmark.com/leaderboard}}, we decided to use BERT~\cite{Devlin:18} as the base model for our work.
An \textit{uncased}, \textit{base} BERT architecture was used as the starting checkpoint and fine-tuned on the generated training data.
The source document and claim sentence were fed as input to the model and the two-way classification (\PositiveLabel{}/\NegativeLabel{}) was done using a single-layer classifier based on the \texttt{[CLS]} token.
We refer to this model as the factual consistency checking model (\textbf{\BaseModel{}}).

We also trained a version of \BaseModel{} with additional span selection heads using supervision of start and end indices for selection and transformation spans in the source and claim.
The span selection heads allow the  model not only to classify the consistency of the claim, but also highlight spans in the source document that contain the support for the claim and spans in the claim where a possible mistake was made.
We refer to this model as the factual consistency checking model  with explanations (\textbf{\ExplainableModel{}}).

\ModelPerformanceTable
\ExplainableOutputsTable
\section{Experiments}
\SentencePairPerformanceTable

\subsection*{Experimental Setup}
Training data was generated as described in Section~\ref{sec:generated-data} using news articles from the CNN/DailyMail dataset~\citep{Nallapati:16} as source documents.
1,003,355 training examples were created, out of which 50.2\% were labeled as negative (\NegativeLabel{}) and the remaining 49.8\% were labeled as positive (\PositiveLabel{}).

Models described in this work were implemented using the Huggingface Transformers library~\citep{Wolf:19} written in PyTorch~\citep{Paszke:17}.
An \textit{uncased}, \textit{base} BERT model pre-trained on English data was used as the starting point for all experiments.
Models were trained on the artificially created data for 10 epochs using batch size of 12 examples and learning rate of \texttt{2e-5}.
Best model checkpoints were chosen based on the performance on the validation set, final model performance was evaluated on the test set, both described in Section~\ref{sec:manual-data}.
Experiments were conducted using 8 Nvidia V100 GPUs with 16GB of memory.

\subsection*{Results}
To verify how other datasets, from related tasks, transfer to the task of verifying factual correctness of summarization models we trained fact consistency checking models on the MNLI entailment data~\citep{Williams:18} and FEVER fact-checking data~\citep{Thorne:18}.
For fair comparison, before training, we removed examples assigned to the neutral class from both of the datasets.
Table~\ref{tab:model-performance} shows the performance of trained models evaluated by means of class-balanced accuracy and F1 score.
Results show that our \BaseModel{} model substantially outperforms classifiers trained on the MNLI and FEVER datasets, despite being trained using weakly-supervised data.
The performance differences between models can be explained by a domain gap between the examples in MNLI and FEVER and news articles in CNN/DailyMail.
%
It is also likely the errors made by neural summarization models are specific enough not to be present in any of the other datasets, especially those where examples were obtained from human annotators.

To compare our model with other NLI models for factual consistency checking, we conducted the sentence ranking experiment described by \citet{Falke:19} using the test data provided by the authors.
In this experiment an article sentence is paired with two claim sentences, \textit{positive} and \textit{negative}.
The goal is to see how often a model assigns a higher probability of being correct to the positive rather than the negative claim.
Results are presented in Table~\ref{tab:sentence-pair-performance}.
Despite being trained in a \textit{(document, sentence)} setting, our model transfers well to the \textit{(sentence-sentence} setting and outperforms all other NLI models, including BERT fine-tuned on the MNLI dataset.
We were unable to recreate the summary re-ranking experiment because the test data was not made publicly available.

Table~\ref{tab:model-performance} also shows the performance of our explainable model, \ExplainableModel{}.
Metrics show a small drop of performance in comparison to the classifier-only \BaseModel{}, however the explainable model still substantially outperforms the other two models while also returning informative span selections.
Examples of span selections generated by \ExplainableModel{} are show in Table~\ref{tab:explainable-predictions}.
The test set consists of model-generated summaries that do not have annotations for quantifying the quality of spans returned by \ExplainableModel{}. 
Instead, span quality is measured through human evaluation and discussed in Section~\ref{sec:analysis}.

\section{Analysis}
\label{sec:analysis}
\MturkResultsTable

To further inspect the performance of proposed models, we conducted a series of human-based experiments and manually inspected the outputs of the models.

\subsection{Crowdsourced Experiments}
Experiments using human annotators on the MTurk platform demonstrated that the span highlights returned by \ExplainableModel{} are useful tools for researchers and crowdsource workers manually assessing the factual consistency of summaries.
For each experiment, examples were annotated by 3 human judges selected from English-speaking countries.
Annotator compensation was set to ensure a 10 USD hourly rate.
These experiments used 100 examples sampled from the manually annotated test set.
Data points were sampled to ensure an equal split between \PositiveLabel{} and \NegativeLabel{}  examples.

To establish whether model generated spans in the article and claim are helpful for the task of fact checking, we hired human annotators to complete the mentioned task.
Each of the presented \textit{document-sentence} was augmented with the highlighted spans output by \ExplainableModel{}.
Judges were asked to evaluate the correctness of the claim and instructed to use the provided segment highlights only as suggestions.
After the annotation task, judges where asked whether they found the highlighted spans helpful for solving the task.
Helpfulness of \textit{article} and \textit{claim} highlights was evaluated separately.
The left part of Table~\ref{tab:mturk-results} presents the results of the survey.
A combined number of $91.75\%$ annotators found the article highlights at least somewhat helpful for the task, $81.33\%$ of annotators declared the claim highlights as at least somewhat helpful.
To verify whether low quality judges do not bias the presented scores, we applied different data filters to the annotations:
\textit{Raw Data} considered all submitted annotations, \textit{Golden Aligned} only considered annotations where the annotator-assigned label aligned with the author-assigned label for the example, \textit{Majority Aligned} only considered examples where the annotator-assigned aligned with the majority-vote label assigned for the example by all judges.
As shown in Table~\ref{tab:mturk-results}, filtering the annotations does not yield substantially changes in the helpfulness assessment.

Despite instructing the annotators to consider the provided highlights only as a suggestion when solving the underlying task, the annotators perception of the task could have been biased by the model-highlighted spans.
To check how well the generated spans align with an unbiased human judgement, we repeated the previous experiment with the difference that model generated highlights were not displayed to the annotators.
Contrarily, the annotators were asked to solve the underlying task, and highlight the spans of the source and claim that they found adequate.
Using the annotations provided by the a judges we computed the overlap between the model generated spans and unbiased human spans.
Results are shown in the right part of Table~\ref{tab:mturk-results}.
The overlap between spans was evaluated using two metrics - \textit{accuracy} based on a binary score whether the entire model-generated span was contained within the human selected span and \textit{F1} score between the tokens of the two spans, with human selected spans were considered ground-truth.
Results show $65.33\%$ and $65.66\%$ \textit{accuracy}, and $0.6207$ and $0.6650$ \textit{F1} for the \textit{article} and \textit{claim} highlights accordingly.
Similarly to the previous experiment, we applied different data filters to check how the quality of annotations affects the score and found that removing noisy annotations increases both \textit{accuracy} and the \textit{F1} score.

To verify that providing highlights to annotators has positive effect on the efficiency of annotations we ran two factual consistency annotation tasks in parallel, where in one highlights were provided to the annotators and the other did not show highlights.
We measured the effects of providing highlights on the average time spent by an annotator on the task and the inter-annotator agreement of annotations.
Results are shown in Table~\ref{tab:highlight-improvements}.
The experiment showed that when completing the task with highlights, annotators were able to complete it $21\%$ faster and the inter-annotator agreement, measured with Fleiss' $\kappa$, increased by $38\%$.

Results obtained through crowdsourcing tasks support the hypothesis that the span selections generated by our explainable model can be a valuable asset for supporting human-based factual consistency checking.

\subsection{Limitations}\label{sec:shortcomings}
In order to better understand the limitations of the proposed approach, we manually inspected examples that were misclassified by our models.
The majority of error made by our fact checking model were related to commonsense mistakes made by summarization models.
Such errors are easy to spot for humans, but hard to define as a set of transformations that would allow such errors to be added to the training data.

In addition, certain types of errors stemming from dependencies between different sentences within the summary, such as temporal inconsistencies or incorrect coreference, are not handled by the \textit{document-sentence} setting used in this work. 

\HighlightImprovementsTable{}
\section{Conclusions}
We introduced a novel approach to verifying the factual consistency of summaries generated by abstractive neural models.
In our approach models are trained to perform factual consistency checking on the \textit{document-sentence} level that allows them to handle a broader range of errors in comparison to the previously proposed \textit{sentence-sentence} approaches.
Models are trained using artificially generated, weakly-supervised data created based on insights coming from the analysis of errors made by state-of-the-art summarization models.
Through quantitative studies we showed that the proposed approach outperforms other models trained on textual entailment and fact-checking data.
A series of human-based experiments showed that the proposed approach, including the explainable factual consistency checking model can be a valuable tool for assisting humans checking for factual consistency.

Shortcomings of our approach explained in Section~\ref{sec:shortcomings} can be treated as guidelines for potential future work.
The methods proposed in this work could be expanded with more advanced data augmentation techniques, such as generating claims that cover multi-sentence spans from the source document or include commonsense mistakes.

We hope that this work will encourage more research efforts in the important task of verifying and improving the factual consistency of abstractive summarization models.
\section*{Acknowledgements}
We thank Nitish Shirish Keskar, Dragomir Radev, Ben Krause, and Wenpeng Yin for reviewing this manuscript and providing valuable feedback.

\bibliography{acl2019}
\bibliographystyle{acl_natbib}
\end{document}